%% file: main.tex
\documentclass[conference]{IEEEtran}

\usepackage[style=ieee,sorting=none]{biblatex} 

\usepackage{graphicx}
\usepackage{graphics}
\usepackage{array}
\usepackage{comment}
\usepackage{siunitx}
\usepackage{hyperref}
\usepackage[all]{hypcap} 
\usepackage{tabularx}
\usepackage{colortbl}
\usepackage{tablestyles}
\usepackage{booktabs}
\usepackage{xcolor} 
\usepackage{xfrac}

\newcommand{\ra}[1]{\renewcommand{\arraystretch}{#1}}

\title{\LARGE \bf
Characterization of Semantic Segmentation Models on Mobile Platforms for Self-Navigation in Disaster-Struck Zones}

\begin{document}

\author{\IEEEauthorblockN{Ryan Zelek and Hyeran Jeon}
\IEEEauthorblockA{Computer Science and Engineering Department\\
University of California, Merced\\
Merced, CA, USA\\
ryanzelek@gmail.com, hjeon7@ucmerced.edu}
}

\maketitle
\pagestyle{empty}

\input{abstract}

\input{introduction}
\input{related_work}
\input{earthquake_site_image_database}
\input{platform_structure}

\input{characterization}
\input{optimizations}
\input{analysis}
\input{conclusion}
\input{bibliography}

\addtolength{\textheight}{-12cm}   








\end{document}

%% file: abstract.tex
\begin{abstract}
The role of unmanned vehicles for searching and localizing the victims in disaster impacted areas such as earthquake-struck zones is getting more important. Self-navigation on an earthquake zone has a unique challenge of detecting irregularly shaped obstacles such as road cracks, debris on the streets, and water puddles. In this paper, we characterize a number of state-of-the-art FCN models on mobile embedded platforms for self-navigation at these sites containing extremely irregular obstacles. We evaluate the models in terms of accuracy, performance, and energy efficiency. We present a few optimizations for our designed vision system. Lastly, we discuss the trade-offs of these models for a couple of mobile platforms that can each perform self-navigation. To enable vehicles to safely navigate earthquake-struck zones, we compiled a new annotated image database  of various earthquake impacted regions 
that is different than traditional road damage databases. We train our database with a number of state-of-the-art semantic segmentation models in order to identify obstacles unique to earthquake-struck zones.  Based on the statistics and tradeoffs, an optimal CNN model is selected for the mobile vehicular platforms, which we apply to both low-power and extremely low-power configurations of our design. To our best knowledge, this is the first study that identifies unique challenges and discusses the accuracy, performance, and energy impact of edge-based self-navigation mobile vehicles for earthquake-struck zones. Our proposed database and trained models are publicly available. 

\end{abstract}

%% file: introduction.tex
\section{INTRODUCTION}

With global-wide monitoring and technology evolution, the riskiness of natural disaster is getting lower. However, there are still deadly incidents such as earthquakes. Earthquakes are a geologic inevitability of some countries and states. 
To reduce the effects from earthquakes, we should think about the ways to quickly search earthquake-struck zones and safely rescue more lives. For this purpose, unmanned autonomous vehicles will be effective, which navigate impacted sites while localizing the people and reporting the damages without risking the lives of others, such as firefighters and other rescue workers. In the natural disaster impacted sites, edge-based small automobiles would be more cost effective and feasible solutions because these small cars can navigate underneath collapsed buildings, unstable bridges, and narrow walkways that are difficult to be driven in by a full-sized vehicle, and also hidden from the view angles of aerial solutions such as drones. Though the self-navigation research and industry has been quickly evolving, most of the solutions focus on detecting fairly regular-shaped objects such as humans, buildings, and streets from a full-sized vehicle’s perspective. However, these solutions cannot be directly applied to earthquake-struck zones because of the unique obstacles such as debris, cracks, and puddles on the streets, which the mobile edge vehicle perceives in a completely different perspective.

\begin{figure}[htp]
    \vspace*{0.2cm}
    \centering
    \includegraphics[width=8.6cm]{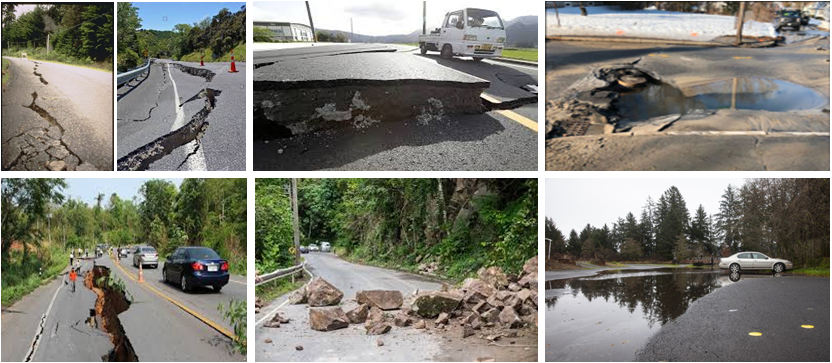}
    \caption{A few images from our proposed Earthquake-site database }
    \label{dataset_imgs}
\end{figure}

In this study, we present an edge-based unmanned vehicle that can navigate disaster-struck sites by avoiding extremely irregular-shaped obstacles. To understand the design tradeoffs, we provide detailed characterizations of recognition accuracy, performance, and energy impact of various system components. We develop obstacle avoidance algorithm with fully convolutional network (FCN) that are known to be effective to recognize irregular-shaped objects. We compare various state-of-the-art FCN models on multiple mobile platforms, which will give insights to the researchers and developers about the pros and cons of different FCN models so that they can choose the right one for their purposes. With these results, we also show a few optimization ideas that speedup the end-to-end execution by 75\%. 
We compile and release a new annotated image database of earthquake-struck sites at \emph{\href{https://gitlab.com/thor-auto1/thor-auto}{https://gitlab.com/thor-auto1/thor-auto}}. 
\begin{table*}[htp]
\vspace*{0.2cm}
\caption{Comparison of the most relevant databases to ours}
\scriptsize
\label{datasets_table}
\begin{center}
\begin{tabular}{m{3.25cm}m{1.5cm}m{3cm}m{3cm}m{2.15cm}m{0.8cm}m{1cm}}
\toprule
Database &
Labeling Method &
Camera Perspective &  
Detectable Classes &
Application &
Total Images &
Crack Severity\\
\midrule

\rowcolor{black!20}
GAPv2 \cite{gapv2} &
Bounding Box &
Perpendicular to ground from top of vehicle height &
Road Cracks &
Automated road \newline inspection &
2468 &
Primarily Minor\\

Pavement Distress Detection \cite{pavementDistress} &
Bounding Box &
Human height, Aerial view about half a meter high &
Road Cracks &
Automated road \newline inspection &
1200 &
Primarily Minor\\

\rowcolor{black!20}
Road Damage Dataset \cite{roadDamageDetector} &
Bounding Box &
Front of full-size vehicle&
Road Cracks &
Automated road \newline inspection &
9053 &
Primarily Minor\\

GAPs384 \cite{gaps384} &
Pixel-level &
Perpendicular to ground from top of vehicle height&
Road Cracks &
Automated road \newline inspection &
384 &
Primarily Minor\\

\rowcolor{black!20}
CRACK500 \cite{crack500}\newline
CrackTree \cite{cracktree}\newline
CrackForest \cite{crackforest}\newline
CrackIT \cite{crackit} &
Pixel-level &
Aerial view about half a meter high &
Road Cracks &
Automated road \newline inspection &
500\newline
206\newline
118\newline
56 &
Primarily Minor\\

CrackU-net \cite{cracku_net} &
Pixel-level &
Front of full-size vehicle and Aerial view about half a meter high &
Road Cracks &
Automated road \newline inspection &
3000 &
Primarily Minor\\

\rowcolor{black!20}
Lost and Found \cite{lostAndFound} &
Pixel-level &
Front of full-size vehicle&
Small Obstacles on Roads &
Self-driving vehicle in regular environments &
2104 &
N/A \\

Thor Database (ours) &
Pixel-level &
Front of full-size and miniature-size vehicle, drone height, human height&
Road Cracks, Vehicles, Buildings, Water Puddles, Humans, Vegetation, Sky, Other Irregular Objects &
Self-driving vehicle in earthquake-impacted environments &
737 &
Primarily Major\\
\bottomrule

\end{tabular}
\end{center}
\end{table*}

The main contributions of this work are as follows:
\begin{itemize}
    \item We present a proof-of-concept vehicular system design, namely \textit{Thor} that can navigate disaster-struck sites. It consists of mobile CPU-GPU SoC platform that can run the full pipeline of self-navigation. The vehicular platform uses a combination of Lidar and FCN recognition outputs to make decisions for collision avoidance at the fields that have extremely irregular-shape obstacles. 
    \item Thorough evaluations and in-depth comparisons are conducted for six state-of-the-art FCN models in the perspectives of accuracy, performance, and energy-efficiency on two mobile CPU-GPU SoC platforms. To the best of our knowledge this is the first study that shows in-depth evaluations of FCN models on embedded platforms. 
    \item We present various optimizations to tackle inefficiencies observed from the evaluations of FCN models on embedded platforms.  
    \item We develop a new earthquake-site image database that is annotated with pixel-level labels for semantic segmentation based detection algorithms. Our database contains various obstacles that reside at earthquake-struck sites unlike existing databases that are collections of road cracks for road maintenance purposes. 
    
\end{itemize}

%% file: related_work.tex
\section{RELATED WORK and MOTIVATION}

Small unmanned vehicles have a lot of potential to be used to rescue people when disaster occurs. However, there has not been many studies done for developing image database or model evaluations for the disaster-struck sites such as earthquake. Given that most of the obstacle detection algorithms use supervised learning, it is impossible to develop navigation algorithms without a training dataset that reflects the unique environment of the target sites. Though myriad of image recognition algorithms have been developed for self-navigating vehicles, most of them are designed for relatively regular-shaped obstacles such as building, human, dogs, road signs and so on. Therefore, it is still unclear if the existing technology is good enough to be applied for the area that has unique environment. We aim to show the case study for earthquake-struck sites where the visual field will be filled with cracks, broken bridge, debris, unclear street boarders and so on. 

In an earthquake-struck zone, the most obvious hazard to a vehicle is severely damaged road. There are a number of such databases that can be used to detect road damage and various obstacles on the road, but most of them are designed for road maintenance and hence none can be used for autonomous navigation in earthquake-struck zones, as summarized in Table \ref{datasets_table}. Road damage detection has been thoroughly studied over the years, but where the end goal was to automate the inspection process and plan the regular maintenance repair \cite{maeda_smartphone, gapv2, gaps384, crack500, cracktree, crackforest, pavementDistress, crackit, roadDamageDetector, cracku_net}. This application difference is important because it defines the problem and how image features are represented. From earthquakes, the road damage caused by the seismic waves traveling through the earth is much greater than that of normal road degradation, and requires a new methodology for detecting road damage. 
Most of the current road damage detection solutions aim to detect the smallest of cracks \cite{gapv2, crackforest, pavementDistress, crackit, roadDamageDetector}, whereas in our case, such small of cracks would minimally impact the stability of the self-driving vehicle. We consider these minor-sized cracks as generally within millimeters or centimeters wide, whereas the cracks considered for self-navigation in earthquake-struck zones are major-sized and are generally within tens of centimeters wide or larger.

Furthermore, a majority of the current solutions detect cracks from a perspective that is either at or near perpendicular to the ground. Although the clearest of features can be extracted from that perspective, the autonomous ground vehicle sees them completely differently, and its perspective also needs to be considered. For ground-based vehicles, cameras are typically mounted on the front, and face in the forward direction with only a slight downward angle. This type of perspective keeps a safe distance while enabling perception on the road, obstacles, and background. 

Beyond those listed in Table \ref{datasets_table}, there are other pixel-based databases used for regular vehicle navigation tasks, such as Cityscapes \cite{cityscapes}, KITTI \cite{KITTI}, and CamVid \cite{camvid}. They can detect the regular obstacles such as buildings, vehicles, humans, for a full-sized vehicle environment, but they cannot detect road damage and they don't have the viewing perspective of the low-to-the-ground miniature car. 

%% file: earthquake_site_image_database.tex
\section{EARTHQUAKE SITE IMAGE DATABASE}

\begin{figure}[htp]
    \vspace*{0.2cm}
    \centering
    \includegraphics[width=8.6cm]{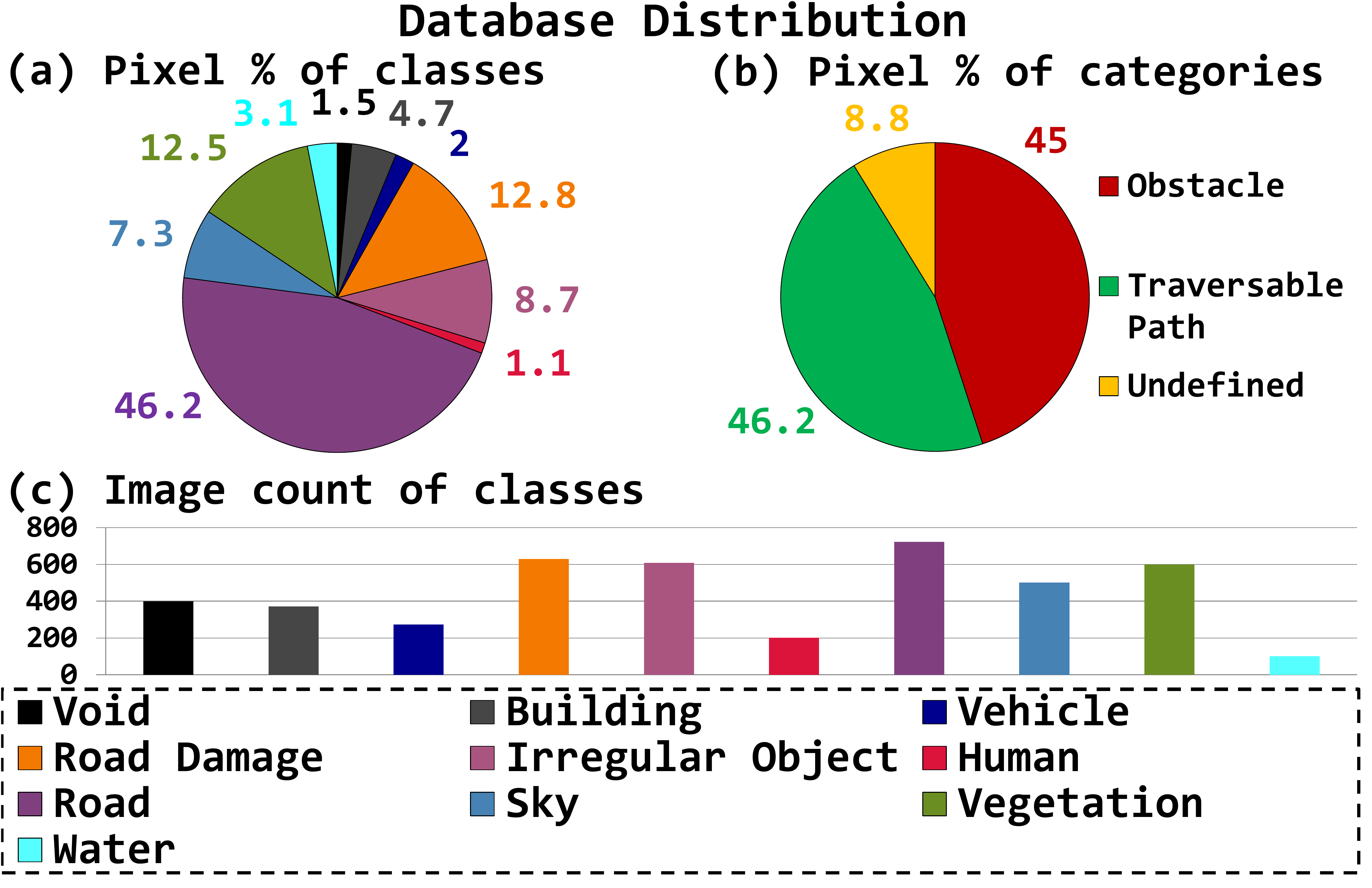}
    \caption{Database distribution in terms of: (a) Per-class at a pixel level, in percentage (b) Per-category at a pixel-level, in percentage (c) The number of images where classes are present, out of 737 total images.}
    \label{dataset_distribution}
\end{figure}

As there is not a publicly accessible earthquake site image database, we compiled a new image database with 737 images that contain objects and scenery that can be found from earthquake-struck sites. The images mainly include classifications of road cracks, water puddles, vehicles, humans, road, sky, vegetation, and other irregular objects such as debris. 
Of those classes, their categories are broken down into three main types. \textbf{Obstacles}, which includes objects that the vehicle should avoid driving into. \textbf{Traversable path}, where it is safe for the vehicle to drive through. Lastly, the \textbf{undefined} category refers to those classes that do not play a role in obstacle avoidance, which in this work, we consider as the sky and void label types. Void labeled pixels, such as blurry pixels in the far background of some images, are unclear even by human eyes, and such pixels cannot be truthfully labeled with a classification. These void labeled pixels are not predicted by our system and are disregarded from our loss function when training the convolutional neural network (CNN) models. Figure \ref{dataset_distribution} shows the distribution of the classes and categories in our database.

The environment of data collection includes both non-shadowy and partially shadowy environments where direct sunlight may not occur. 
To achieve a high detection accuracy on edge-based vehicles where the viewing angle from the edge devices is much lower than that of full-size vehicles or drones, we considered right angles that can be seen by the vehicle at the field drive while collecting images. 
We annotated the irregular objects with ground truth labels using PixelAnnotationTool \cite{pixel_annotation_tool}. Our database is used for semantic segmentation and training fully convolutional neural network (FCN) models. 

%% file: platform_structure.tex
\section{Platform Architecture}
\subsection{Vehicle Platform Configuration}

\begin{figure}[htp]
    \centering
    \includegraphics[width=8.6cm]{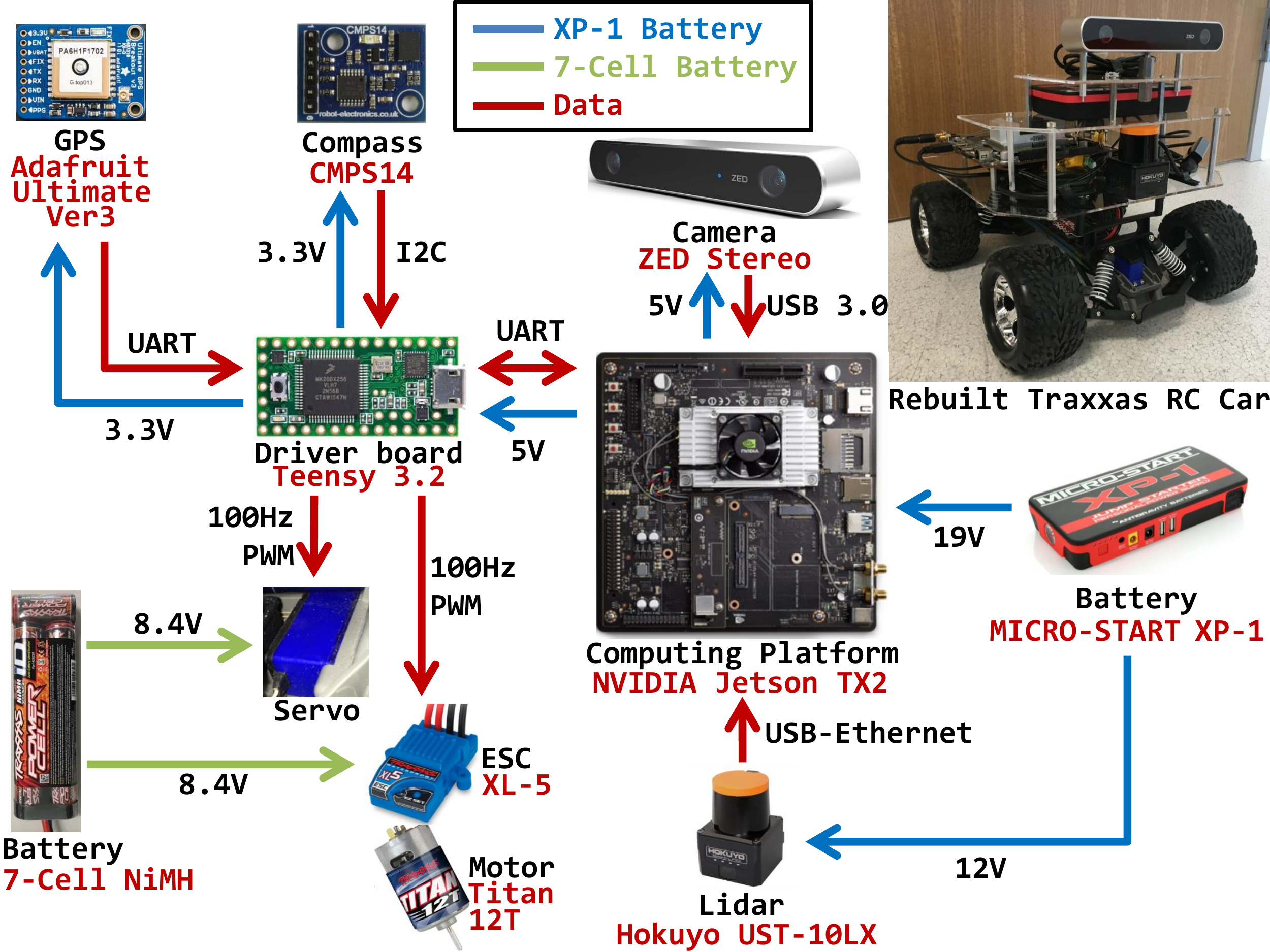}
    \caption{Hardware design of the proposed system}
    \label{hw_breakdown}
\end{figure}

Figure \ref{hw_breakdown} shows the hardware design of the developed edge-based self-navigation vehicle.
We tested NVIDIA Jetson TX2 \cite{tx2} and Xavier AGX \cite{agx} computing platforms (in MAX-N mode), which are interchangable in our design. The TX2 platform consists of a CPU cluster of a Denver2 dual-core CPU and an ARM Cortex A57 quad-core CPU, while the GPU is Pascal-based with 256 cores. The AGX Xavier consists of an 8-core ARM v8.2 64-bit CPU and a 512-core Volta-based GPU. A sensor fusion of a Hokuyo LiDAR range-finder and a ZED stereo camera were incorporated for the vehicle to perceive the environment at a local level. An Adafruit GPS and an CMPS14 Compass globalize the vehicle. The ZED camera input is fed to the GPU for pixel-level object detection. Robotic Operating System (ROS) Melodic is used as the framework for controlling the various components equipped on the car platform.

\subsection{Object Detection and Navigation Pipeline}

\begin{figure}[htp]
    \centering
    \includegraphics[width=8.6cm]{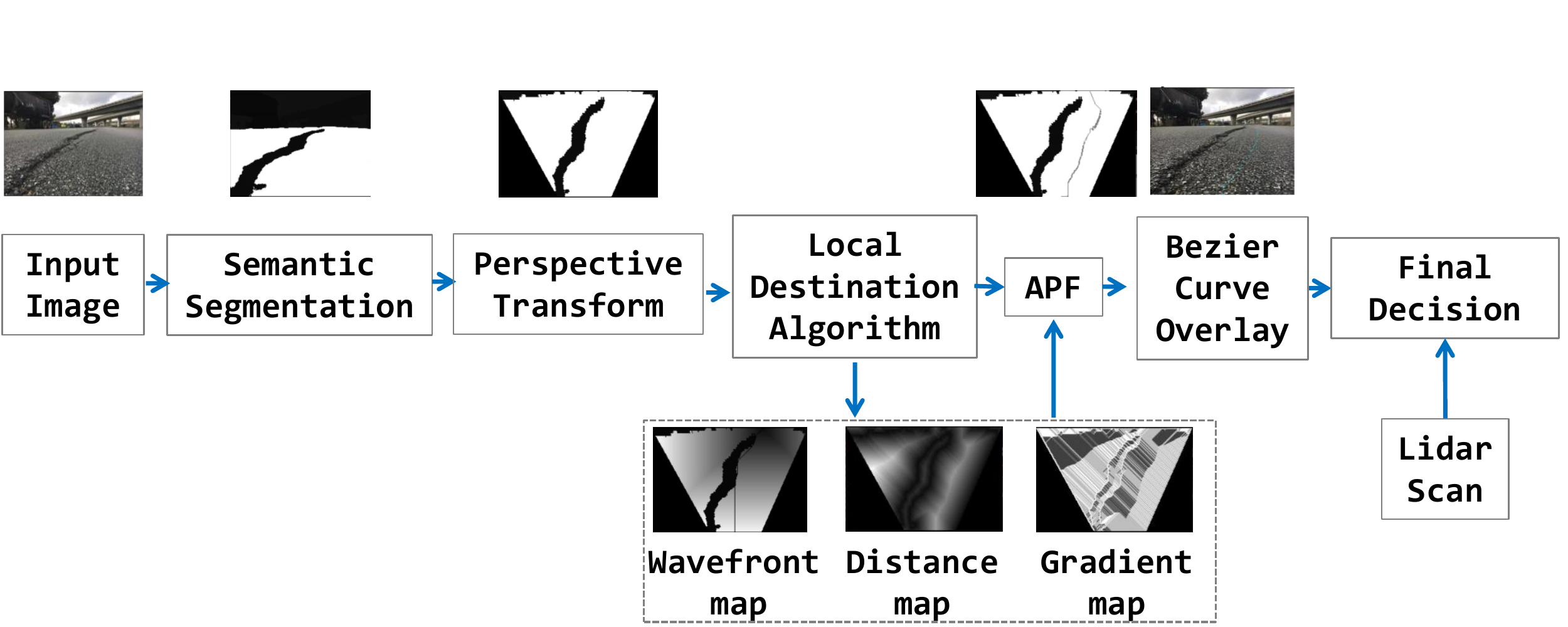}
    \caption{Local navigation software flow}
    \label{sw_structure}
\end{figure}

Our vision system relies upon the camera input and Lidar sensor to detect objects at a local level.
Fig. \ref{sw_structure} shows our software flow for the local navigation task.

Once an image is taken by the camera, the objects are recognized by an FCN via semantic segmentation. 
Then, the object classes and location information as well as Lidar input are used by the navigation algorithm that finalizes the direction and speed of the driving. The Lidar input 
is used to ensure the vehicle steers in the safe direction and in a timely manner. 

The semantic segmentation NN task is executed on the GPU while using the Caffe framework. Because this NN processing has the longest latency and is the bottleneck of our design, we offload the remaining components of our navigation algorithm to be executed on the CPU. 

In terms of these CPU-based steps taken after the semantic segmentation, we initially determine an aerial view by applying a perspective transform to the segmented mask, which is then condensed in half to allow real-time operation.

\input{local_destination}
\input{path_to_the_local_destination}



\input{semantic_segmentation_model_evaluation}

%% file: local_destination.tex
\subsubsection{Local Destination}

From the condensed map, we first determine a safe destination in terms of a pixel-level coordinate. Inspired by the approach used in \cite{small_obstacle}, rows are scanned from the bottom of the mask towards the top. We check all horizontal road intervals that are wide enough for the vehicle to pass through. The center of the largest interval on that row is marked as the local destination. When a row is encountered that has no road interval that’s wide enough to fit the vehicle, the row scanning stage is complete.

%% file: path_to_the_local_destination.tex
\subsubsection{Path to the Local Destination}

After a destination is chosen, a safe path needs to be determined that the vehicle can follow. The path we determine is based on an Artificial Potential Field (APF) \cite{apf} approach. During APF, the vehicle experiences two potentials, an attractive and repulsive. The attractive pushes the vehicle towards the destination, while the repulsive pushes the vehicle away from obstacles.
The attractive force vector ($F_{att}$) is based on the Euclidean distance from the vehicle ($q$) to the local destination ($q_{goal}$), where those two points are defined in terms of pixel-level coordinates:
\begin{equation}
\label{fatt_eq}
F_{att}(q) = -k_{att} \cdot{(q - q_{goal})}
\end{equation}

To navigate around the irregular shaped obstacles in earthquake-struck zones, we consider the repulsive force vector ($F_{rep}$) as an accumulation from eight different vectors surrounding the vehicle. These angles ($\theta_{i}$) consist of all \ang{45} combinations ranging from 0 to 315 degrees. Their magnitude is based on the pixel in that direction's distance to the nearest obstacle ($d_{obst_{i}}$), where a distance map is computed by a common Brushfire algorithm. The term $d_{0}$ is directly related to the projected distance away from the vehicle's current location. If there are no obstacles within $d_{0}$ from the projected location, then the obstacle's repulsive force does not impact that point of the trajectory. As a result, the repulsive force vector ($F_{rep}$) can be determined with the following equation:
\begin{equation}
\label{frep_eq}
F_{rep}(q) = \sum_{i=1}^{8}k_{rep} \cdot (\frac{1}{d_{obst_{i}}(q)} - \frac{1}{d_{0}}) \cdot \frac{\hat{u}(\theta_{i})}{d_{obst_{i}}^2(q)}
\end{equation}
While the net force vector ($F_{net}$) is the sum of the attractive and repulsive force vectors:
\begin{equation}
\label{fnet_eq}
F_{net}(q) = F_{att}(q) + F_{rep}(q)
\end{equation}

We apply gradient descent to the net force to determine a full trajectory. When obstacles are equally distant from the vehicle or when an obstacle is within a threshold too close, then an alternative repulsive force is applied. This force is based on the gradient map, which provides a one-to-one correlation for the direction in which the nearest obstacle can be found, per pixel. The gradient map is computed by iterating through the distance map and determined based on the minimum value of neighboring pixels.
There are a few minor cases where the resulting trajectory cannot complete due to the destination being far away while also being very close to an obstacle, which typically occurs near the upper boundary of the mask.
In those cases, we use data from Wavefront algorithm to complete the trajectory.
There may still be oscillations due to the nature of APF; to smooth the path, we use four points from this trajectory as control points for a cubic Bezier curve. The resulting curve is smooth enough such that the vehicle can feasibly follow it. 

%% file: semantic_segmentation_model_evaluation.tex
\subsection{Semantic Segmentation Models}

To understand the impact of different model architectures, we evaluated six popular FCN models, 
which are:


\begin{itemize}
    \item \textbf{FCN-VGG16} is proposed by Long, Shelhammer, and Darrell \cite{fcn} and has shown effective performance for object classification tasks across generic types of applications. Designed with 16 conv. layers and the heavy encoder from VGG16-net, the output is decoded with a pixel stride width of 32. One of the most commonly trained databases for this network is Pascal VOC database \cite{pascal_voc}, which detects animals, furniture, humans, amongst other basic object types from a human's perspective. 
    
    \item \textbf{FCN8} is also proposed by \cite{fcn} and is an improved version of FCN-VGG16 in that the decoder output stride is taken at eight pixels wide instead of 32. Although this provides additional refinement to the output mask, this is at the cost of requiring additional layers for the decoder.
    
    \item \textbf{FCN-AlexNet} is again proposed by \cite{fcn} from their work on converting classical object detection CNNs to FCN networks. Of which, they converted the lightweight eight conv. layer AlexNet model to a fully convolutional model for semantic segmentation.
    
    \item \textbf{ERFNet} is proposed by Romera et al. \cite{erfnet} and was originally designed for the semantic segmentation task for intelligent vehicle types of applications. ERFNet consists of 23 layers but uses dilated convolutions and residual layers as a couple of ways to keep a low memory footprint. 
    
    \item \textbf{ENet} is proposed by Paszke et al. \cite{enet} and they designed their FCN network specifically for mobile types of applications, including IoT devices and self-driving cars where computing resources are limited. ENet consists of 29 layers and a few of their efficiencies are gained by the use of decomposed convolutions, such as the dilated and asymmetric convolutions they use in their bottleneck-based architecture.  
    
    \item \textbf{SegNet} is proposed by Badrinarayanan et al. \cite{segnet} and is effective for scene-understanding types of applications, as opposed to generic object detection tasks where background information would not have as much consideration. The 27 conv. layer SegNet model also uses a VGG16-based encoder, however, they improve efficiency by their decoder design, which performs upsampling by reusing max-pooling indices from the encoder.
    
\end{itemize}

We developed all these models in Caffe framework~\cite{caffe} 
such that each network could be fairly compared to each other by using the same system configuration and software framework. To accelerate model training and improve the accuracy of regular objects such as vehicles, buildings, and humans, we used transfer learning, where each network was initially trained on larger databases (baseline model) and then fine-tuned with our earthquake-site image database. 

Hyperparameter selection was determined based on the parameters used by the baseline models 
and then optimized to achieve the best prediction accuracy with 
our earthquake image database. Table \ref{hyper_table} shows the hyperparameters that derive the best accuracy for each model. We used 
CUDA 11, cuDNN 7.6, and an NVIDIA TITAN RTX GPU for training. 
As a loss function for training, we incorporate a SoftMax with loss layer and normalize it across the batch size. The models were trained until the losses no longer decreased and the accuracies were stable at an optimum. Images were rotated along the vertical axis to augment the data and further improve generalization.

\begin{table}[htp]
\vspace*{0.2cm}
\caption{Hyperparameters used when training each model}
\ra{1.3}
\scriptsize 
\label{hyper_table}
\begin{center}
\begin{tabular}{@{}m{1.75cm}@{}m{0.65cm}m{0.65cm}m{0.65cm}m{0cm}m{0.65cm}m{0.65cm}m{0.65cm}}
\toprule
 &
FCN8 & 
FCN-VGG16 &
FCN-AlexNet &
 &
ERFNet &
ENet &
SegNet \\
\midrule

Iterations &
240,450 &
480,900 &
240,450 &
 &
58,182 &
164,193 &
126,474 \\

Epoch &
350 &
700 &
350 &
 &
1,016 &
956 &
920 \\

Learning Rate &
1e-4 &
1e-4 &
1e-3 &
 &
5e-4 &
1e-4 &
1e-3 \\

Batch Size &
1 &
1 &
1 &
 &
12 &
4 &
5 \\

Weight Decay &
5e-4 &
5e-4 &
5e-4 &
 &
2e-4 &
1e-4 &
5e-4 \\

Solver &
SGD &
SGD &
SGD &
 &
Adam &
Adam &
SGD \\

Momentum(2) &
0.9 &
0.9 &
0.9 &
 &
0.9 (0.999) &
0.9 (0.999) &
0.9 \\

\cmidrule{2-4} \cmidrule{6-8}
Pretrained Database &
\multicolumn{3}{c}{Pascal VOC \cite{pascal_voc}} &
 &
\multicolumn{3}{c}{Cityscapes \cite{cityscapes}} \\
\bottomrule
\end{tabular}
\end{center}
\end{table}

%% file: characterization.tex
\section{Characterization and Analysis} 

\input{accuracy}

\input{performance}

\input{energy_efficiency}

%% file: accuracy.tex
\subsection{Accuracy}

\begin{figure}[htp]
    \centering
    \includegraphics[width=8.6cm]{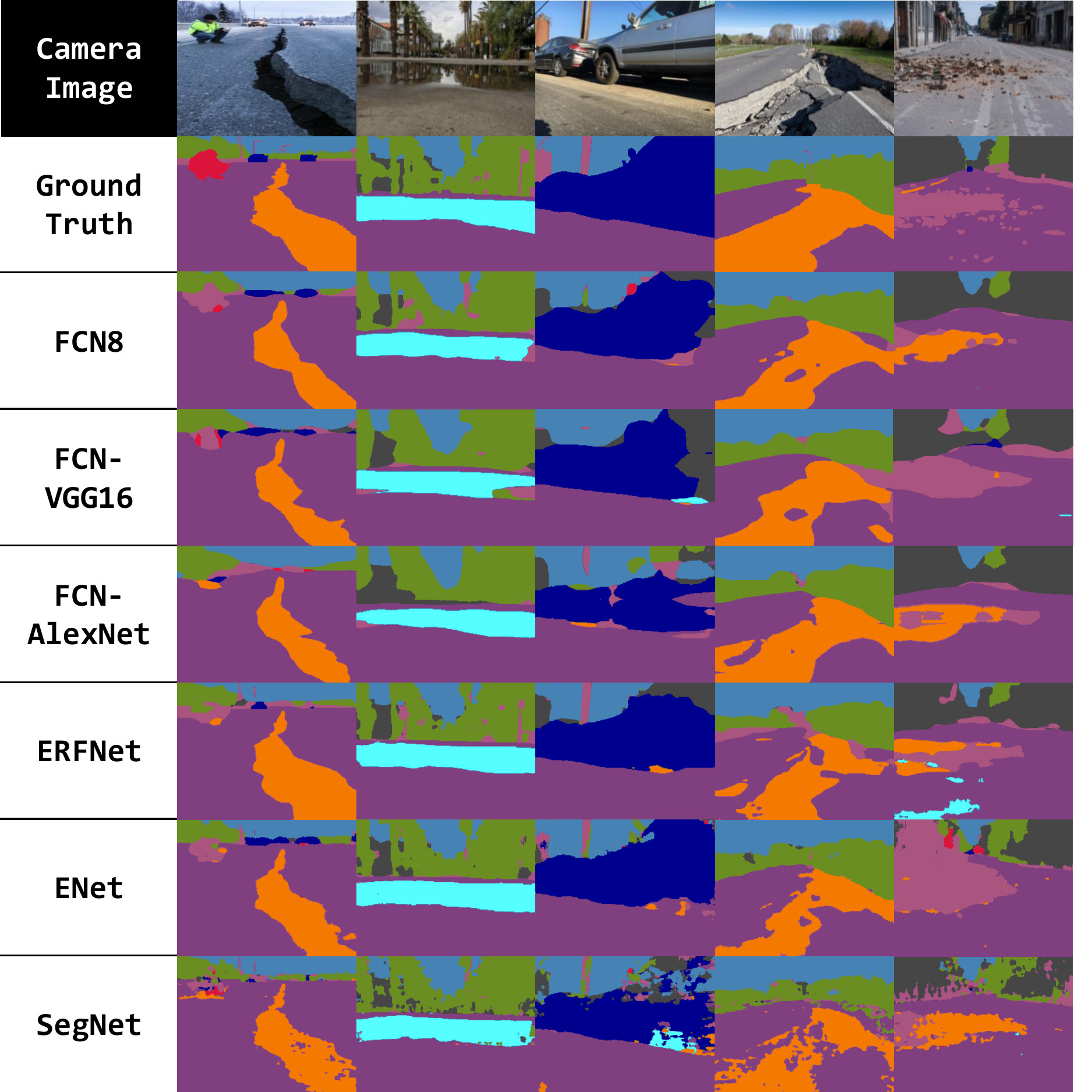}
    \caption{Visualization examples of image segmentation used as a method for qualitatively comparing and evaluating the accuracy of the models.}
    \label{visual_segmentation_comparisons}
\end{figure}

We first evaluated prediction accuracy of the six FCN models with 50 testing images that encompass all of the object classes in environments with both a high and low number of classes present. Figure \ref{visual_segmentation_comparisons} shows a few visualized segmentation results of each network for the earthquake-site database. 
The top row shows the image inputs that include various obstacles such as cracks (the first and the fourth columns), another vehicle (the third column), and debris (the fifth column). The second row shows the ground truth where cracks, other vehicles, and debris are indicated with orange, blue, and light violet colors, respectively. The safe regions for driving are marked with dark violet colored pixels. As can be seen, most of the critical and large obstacles are well recognized by all networks, while some networks show poor pixel-level coverage even for the large obstacles and mis-classification for extremely irregular obstacles such as debris. For example, in the right-most column of Figure \ref{visual_segmentation_comparisons}, a pile of debris is correctly detected and classified by FCN-VGG16 and ENet. The other networks correctly recognized the regions of debris but mis-classified them as cracks. Likely, the crack region on the fourth image is almost completely recognized by FCN8, FCN-VGG16, and FCN-AlexNet, while notably higher number of pixels in the crack regions are classified as safe road by ERFNet, ENet, and SegNet.  

However, we observed that the pixel-level accuracy does not necessarily lead to a collision with the obstacle. As far as the coarse-grained region of the obstacle is recognized as not-passable obstacle, the navigation algorithm made a detour to avoid collision. 
For example, in the example of debris recognition result, even when some networks recognized the debris as crack, as both are not passable obstacles, the vehicle did not pass over them. Similarly, in the example of the crack recognition result, even when some pixels are classified as safe regions, the car did not fall into the crack because majority of the pixels were correctly classified as crack. 
In image recognition algorithms that use semantic segmentation encounter, so called a \textit{pixel to application accuracy disparity}. In other words, pixel-level recognition accuracy may not be the same with application-level accuracy. To accommodate the accuracy disparity, we measured the accuracy in two levels: pixel-level and object-level. 

\begin{figure}[htp]
    \centering
    \includegraphics[width=8.6cm]{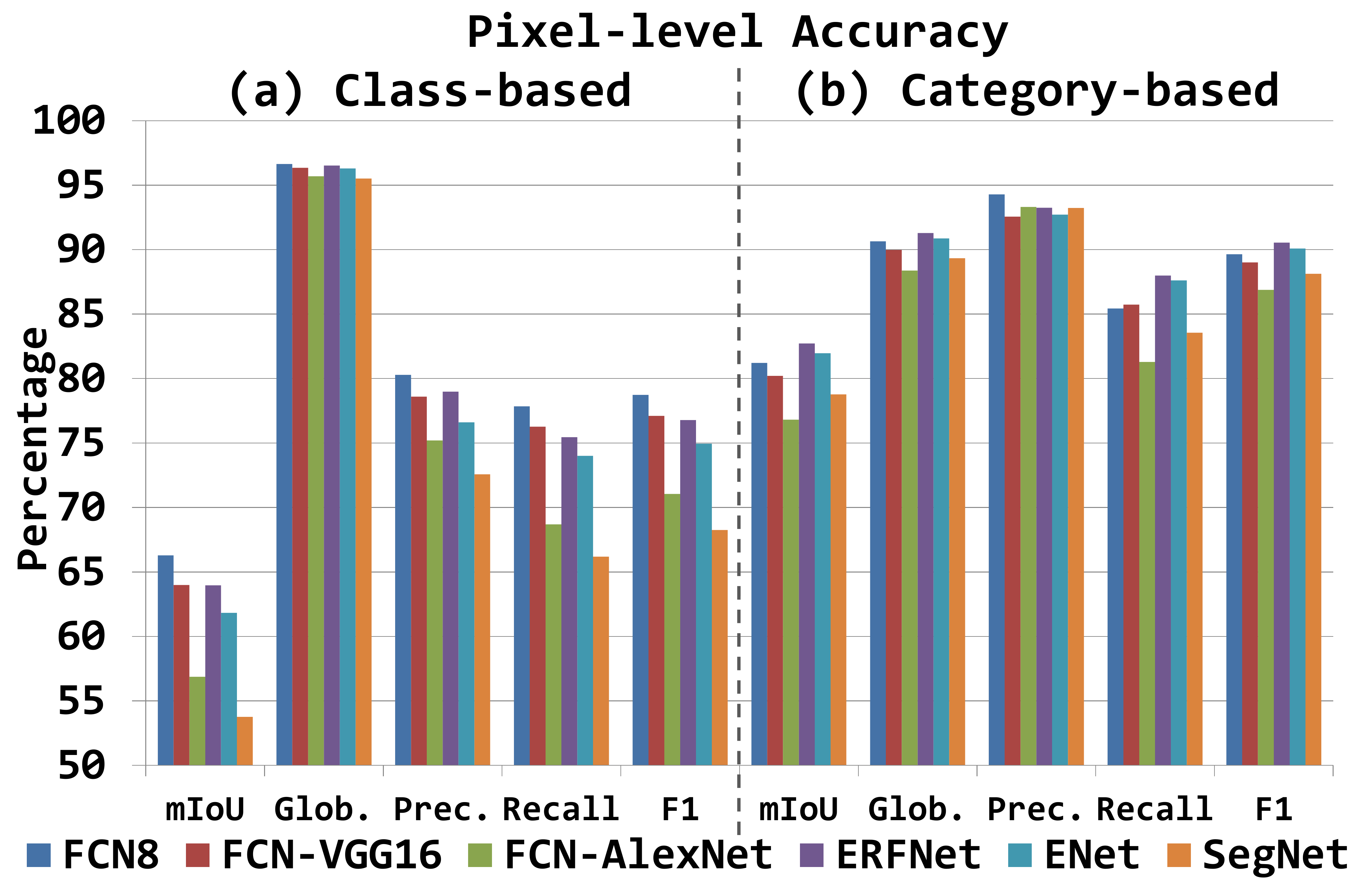}
    \caption{Pixel-level accuracy in terms of (a) Class-based taken from the mean of all classes (b) Category-based taken of obstacles }
    \label{acc_avg_obs}
\end{figure}

Deciding upon the optimal accuracy metric for semantic segmentation models is not a completely solved problem and is still a topic being researched in the community. For example in \cite{new_metric_paper}, they attempt to solve this problem by developing a single accuracy metric that considers both global classification and contour segmentation when defining their proposed singular metric. 
For the pixel-level accuracy evaluations in our work, we use the already well-known accuracy metrics such as IoU, Global, Precision, Recall, and F1 as equated like below, where TP, FP, TN, and FN are true positive, false positive, true negative, and false negative, respectively. With pixel-level accuracy, we can understand how correctly each pixel is classified. The results are shown in Figure~\ref{acc_avg_obs} (a) where IoU takes the average of all classes. The analysis of the results will follow shortly.

\small
\begin{equation}
IoU = \frac{TP}{TP + FP + FN}
\end{equation}
\begin{equation}
Global = \frac{TP + TN}{TP + TN + FP + FN} 
\end{equation}
\begin{equation}
Precision = \frac{TP}{TP + FP}
\end{equation}
\begin{equation}
Recall = \frac{TP}{TP + FN}
\end{equation}
\vspace*{-0.2cm}
\begin{equation}
F1 = \frac{2 \cdot Prec \cdot Recall}{Prec + Recall}
\end{equation}
\normalsize

As noted earlier, pixel-level accuracy does not provide the recognition effectiveness for the navigation. Therefore, we propose to further 
break down the classes into two categories: 
1) the traversable paths and 2) the non-traversable paths that have obstacles that cause collisions. 
To distinguish this from the original pixel-level recognition, we call the original pixel accuracy as \textit{class-based} and this new category accuracy as \textit{category-based} recognition. With category-based recognition, we can understand the prediction accuracy 
that lead to effectiveness of 
the navigation algorithm.


Figure~\ref{acc_avg_obs} shows all five metrics in (a) class-based and (b) category-based recognition. Notable differences were observed in mean IoU where category-based recognition show a significantly higher accuracy than individual pixel class-level recognition. This means that navigation decision will be made with 80\% accuracy even when 40\% of pixels were recognized in incorrect classes. 
Though most of the metrics show higher accuracy in category-based recognition, the Global metric drops about 10\% in category-based recognition. This reduction in global accuracy can be explained by the TN bias associated with the class-based measurement due to its much higher number of classes. Note that TN scores represent pixels where the specific class under test correctly (truly) predicted the pixel as not belong to its class (negative). This means that for images with a large number of classes, the per-class representations would be dominated by the not-belonging (negative) pixels and TN pixels would increase as the number of classes increases. Therefore, lower category-based Global metric is sourced by significantly fewer category-based classes. 

In terms of the other metrics, the precision was around 5-10\% better than the recall in both recognition methods. This precision vs. recall difference implies that the models were slightly better at predicting positive instances of the objects than they were at capturing the entirety of the ground-truth objects. This property benefits scenarios such as self-driving with irregular shaped objects, where the objects need to be detected, but their exact shape and size is not necessarily needed to still be effective for the navigation.


Among the networks, FCN8 consistently showed outstanding accuracy in both recognition methods, while FCN-AlexNet and SegNet were the worst. The highest accuracy with FCN8 is mainly attributed to the largest number of parameters it computes in addition to the low pixel stride it takes at the output (eight pixels). The number of parameters each model computes is shown in Table \ref{num_param_table}. FCN-VGG16 and FCN-AlexNet instead take the output at a larger stride of 32 pixels, which contributes to their lower accuracy. The per-layer parameter breakdown is plotted in Figure~\ref{per_layer_graph}. On the other hand, though ENet and ERFNet stride by only two pixels at the output, they show lower accuracy than larger models because they are originally designed for supporting extremely resource-limited devices with significantly fewer parameters rather than achieving higher accuracy. 
SegNet also computes a larger number of parameters similar to FCN8, but its decoder scheme is completely different from all the other networks. Unlike the other networks that upsample through transposed convolution layers, SegNet does not use transposed convolution layers for upsampling. 
Instead, SegNet reuses pooling indices from the encoder's max pooling layers, which led it to a 
lower detection capability compared to the other models.




\begin{figure}[htp]
    \centering
    \includegraphics[width=8.6cm]{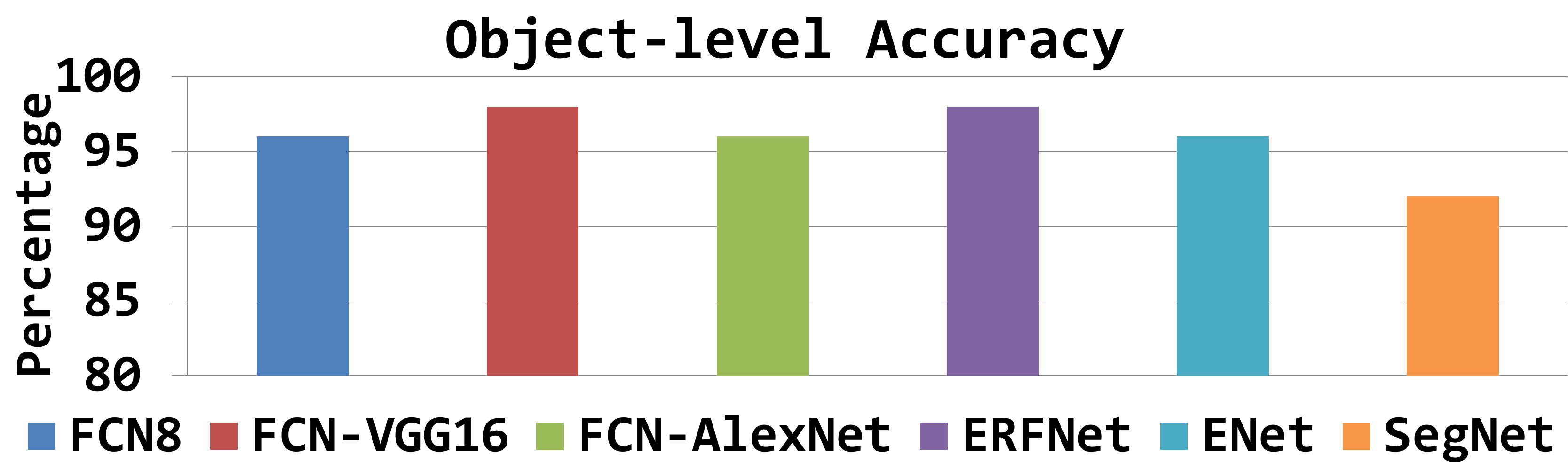}
    \caption{Object-level accuracy of obstacle detection}
    \label{obj_acc_graph}
\end{figure}

In the object-level accuracy, we assume that an object is correctly recognized if the majority of the object region is classified as the target object. In this case, some pixel-level prediction errors are ignored. 
Although the pixel-level metrics are useful for comparing segmentation models, the exact shape and size is not necessarily needed to effectively detect an obstacle, especially for the disaster-struck sites that have extremely irregular-shaped objects. Thus, we also evaluate the detection capability at an object-level, inspired by \cite{detecting_unexpected} and \cite{small_obstacle}, we used 50\% coverage as the threshold to determine object-level detection accuracy. 
These object-level results are shown in Figure \ref{obj_acc_graph}. 

There are a few borderline cases where camera-based vision may fail.
In an earthquake-impacted site, the most relevant case would be where road is broken up in such a way that it has an altered height-angle with respect to the vehicle.
In those cases, the features from the road cracks may not be visible from the models or even human eyes. To handle such cases, our system incorporates a Lidar sensor which can detect the severely elevated or broken up road amongst the other objects that cannot be fully detected by camera-based vision. By using the combined inputs from an FCN model and a Lidar, our car platform showed zero obstacle collision at the field tests.






\textit{\textbf{Observation 1:} Though most of the tested models show above 70\% accuracy in all accuracy evaluation metrics, the models using more parameters and shorter output stride showed better accuracy. }

\textit{\textbf{Observation 2:} Pixel-wise prediction accuracy misleads the navigation accuracy due to the emphasized true negativeness. Category-based prediction accuracy that classifies objects into binary (either traversable or not) can provide more reliable and stable accuracy results for navigation systems. }

%% file: performance.tex
\subsection{Performance}

\begin{figure}[htp]
    \vspace*{0.15cm}
    \centering
    \includegraphics[width=8cm]{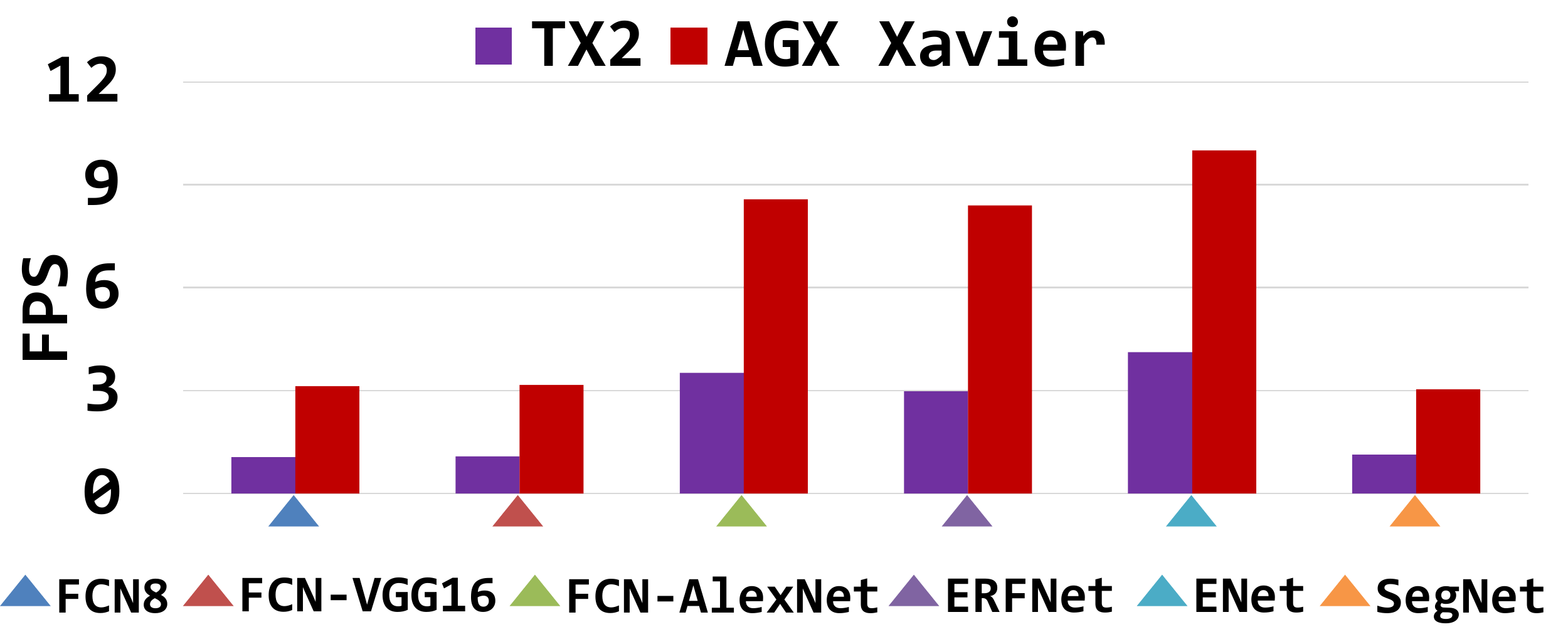}
    \caption{Baseline performance comparison of the FCN models in terms of FPS}
    \label{baseline_fps_graph}
\end{figure}

\begin{figure}[htp]
    \vspace*{0.15cm}
    \centering
    \includegraphics[width=8.6cm]{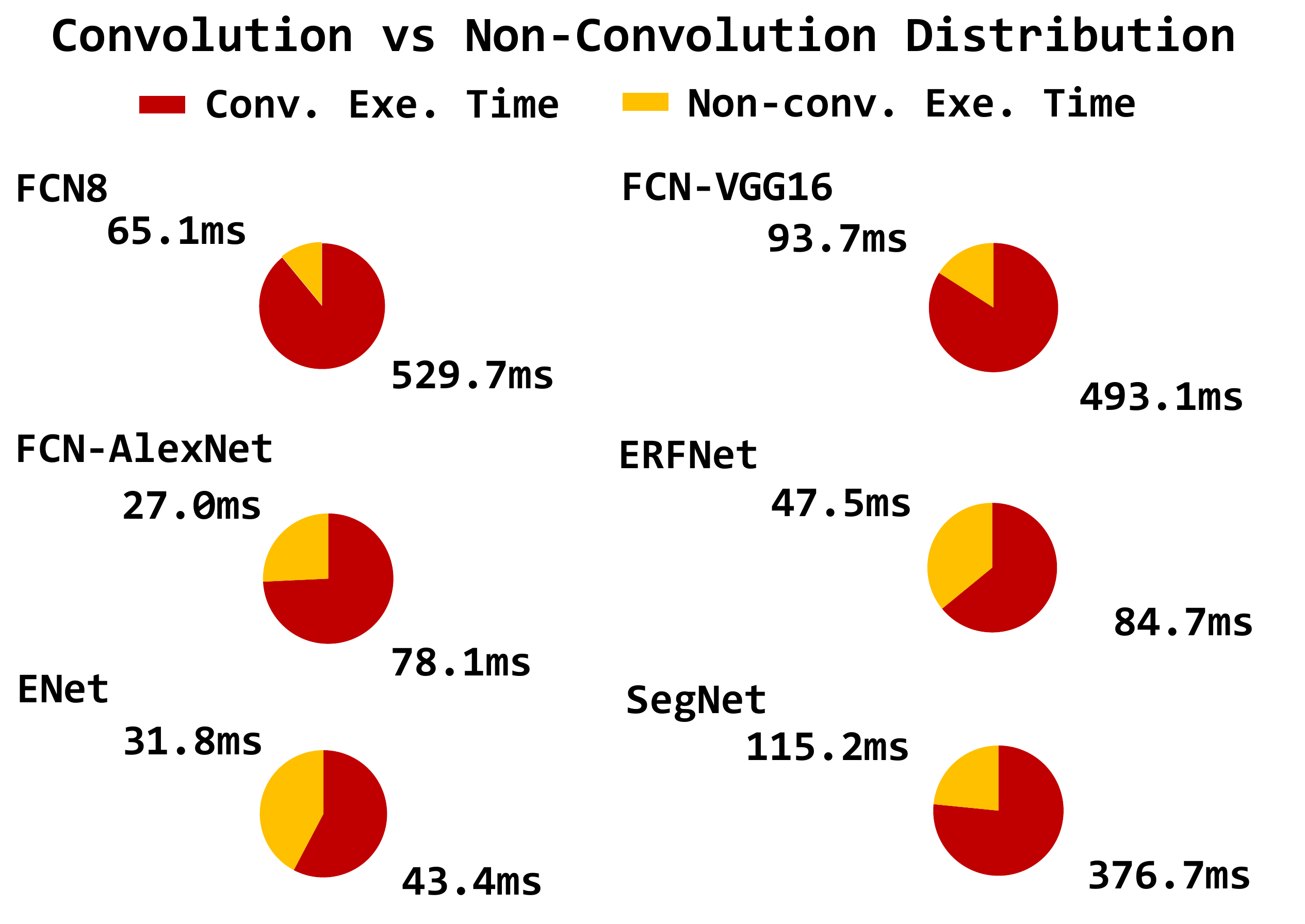}
    \caption{Convolution vs Non-Convolution Distribution }
    \label{conv_dist_graph}
\end{figure}

To recognize obstacles while driving at real-time, the recognition throughput capability in frames per second (FPS) unit is used as an important performance metric in self-navigation systems. The recognition performance is determined by both the efficiency of FCN and the processing hardware. Therefore, we evaluated FPS of the six FCN models on two edge-based GPU platforms that are NVIDIA TX2 and Xavier AGX. Figure~\ref{baseline_fps_graph} shows the results. As can be seen from the figure, Xavier consistently computes with a higher throughput than TX2, which can primarily be attributed to its double the amount of CUDA cores, which stacks up at 512 CUDA cores when compared to the TX2's 256 CUDA cores. These CUDA cores are important because these cores are directly responsible for the convolutional layers in the CNN models. The breakdown of convolution vs non-convolution execution time is shown in Fig. \ref{conv_dist_graph}, which shows that the convolutional layers can dominate the model's computing cost and directly relates to the execution time of the models, and stresses the importance of having a large number of CUDA cores in the platform. The global memory is also important, because that is responsible for holding the weights of these models during execution. TX2 operates with a 128-bit 8GB LPDDR4-3732 memory interface while Xavier operates with a 256-bit 32GB LPDDR4x-4266 memory interface that is both faster and more energy efficient.

\begin{figure}[htp]
    \vspace*{0.15cm}
    \centering
    \includegraphics[width=8.6cm]{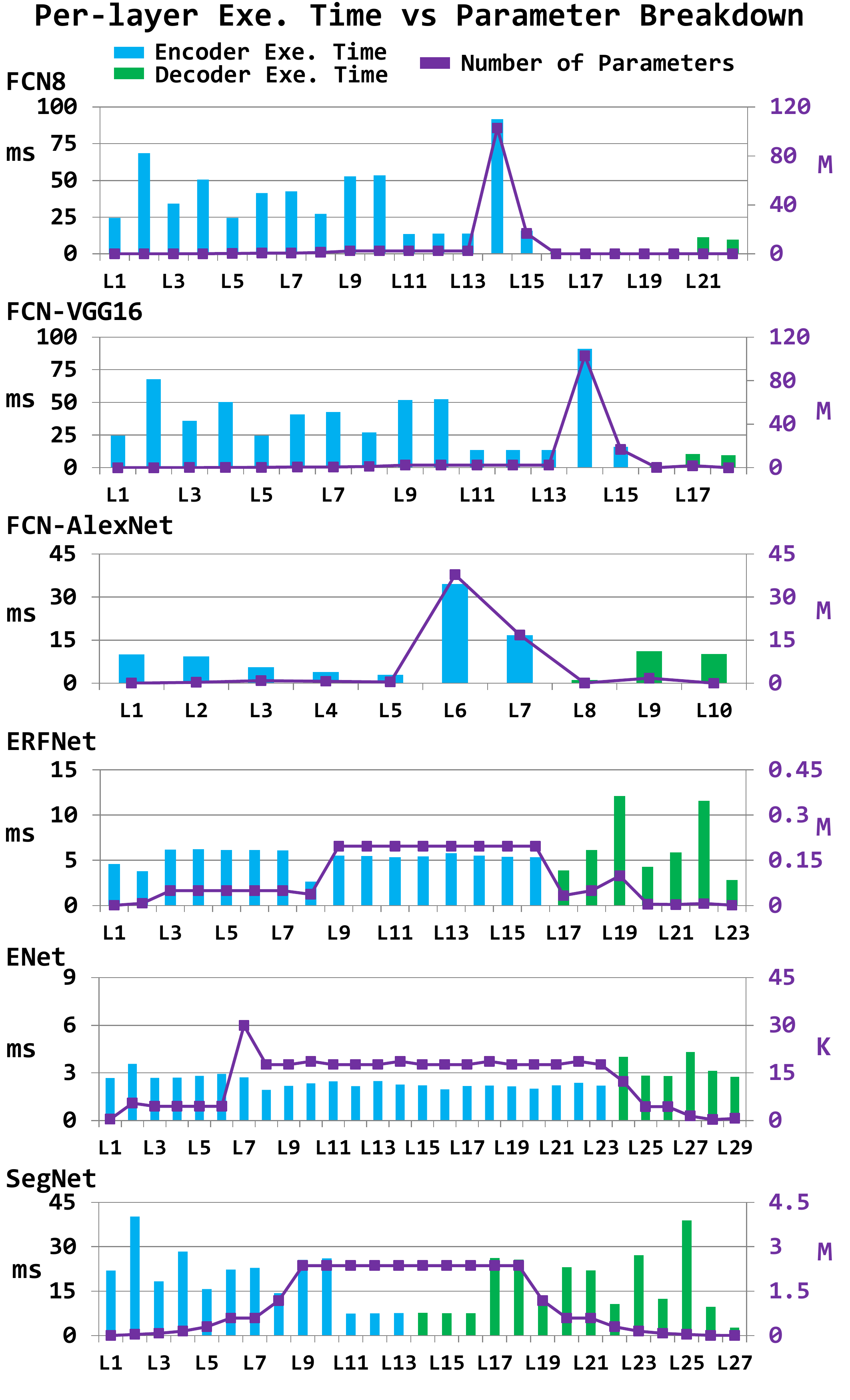}
    \caption{Per-layer Execution time vs Parameter breakdown comparing the networks. In ENet \cite{enet}, these layers are referred to as modules. }
    \label{per_layer_graph}
\end{figure}

Interestingly, FCN-AlexNet, ERFNet, and ENet show 2-3$\times$ better throughput than the others on both platforms. To understand this notable performance difference, we further investigated the per-layer parameter size and latency breakdown as shown in Figure~\ref{per_layer_graph}. As can be seen, one of the notable characteristics of the three models is significantly less parameter usage. As the right-hand side Y-axis and the violet-colored line graphs shows, ERFNet and ENet almost 300\% less parameters than FCN8 and FCN-VGG16. FCN-AlexNet uses more parameter than SegNet but runs almost 30\% fewer layers, and hence layer function invocation latencies are effectively eliminated. Note that each layer needs at least one GPU kernel execution, which also require input and output data transfer. Therefore, FCN-AlexNet achieves good performance by reducing kernel invocation overheads. 


\begin{table}[htp]
\caption{Total Number of Parameters Computed}
\scriptsize 
\label{num_param_table}
\begin{center}
\begin{tabular}{m{1.3cm}m{1.25cm}m{1.1cm}m{0.9cm}m{0.7cm}m{0.9cm}}
\toprule

FCN8 & 
FCN-VGG16 &
FCN-AlexNet &
ERFNet &
ENet &
SegNet \\
\midrule

134,478,048 &
136,140,480 &
58,695,001 &
2,057,503 &
362,563 &
29,432,896 \\

\bottomrule
\end{tabular}
\end{center}
\end{table}

\begin{figure}[htp]
    \vspace*{0.15cm}
    \centering
    \includegraphics[width=8cm]{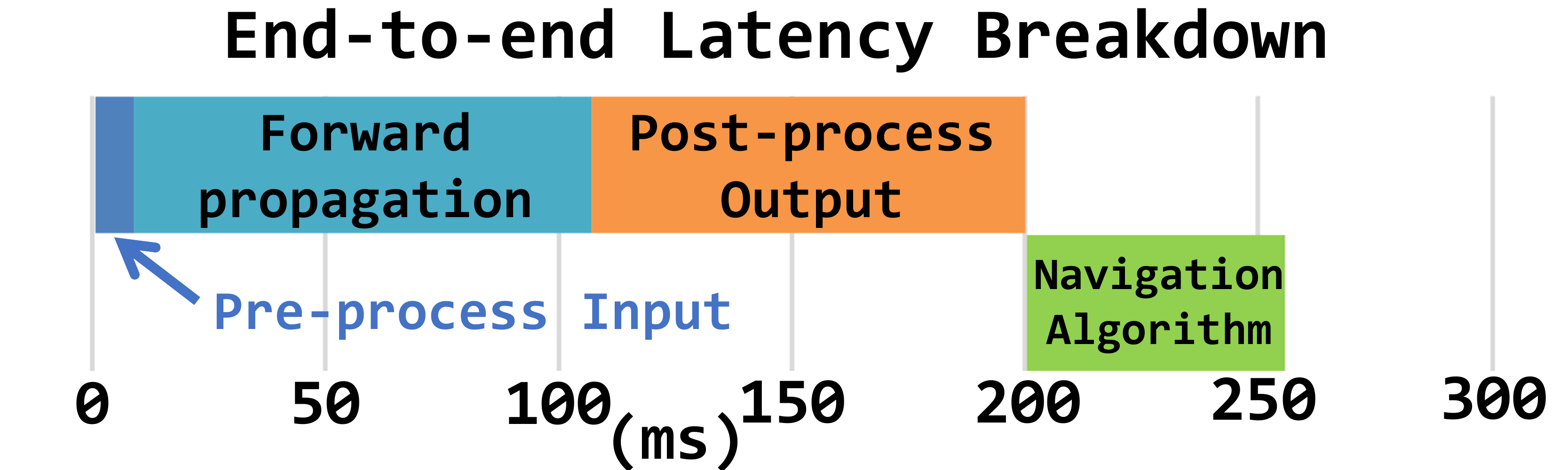}
    \caption{End-to-end latency breakdown (with ENet deployed on the TX2)}
    \label{baseline_latency_graph}
\end{figure}

Next, we breakdown the performance on an end-to-end level, from when a camera frame is received by the ROS node executing the CNN, and until a separate ROS node determines a vehicle direction based on that mask and our navigation algorithm. 
In pre-processing stage, we convert the camera frame to a 32-bit RGB float, then reshape and wrap it onto the input layer of the network.
The forward-propagation delay includes the GPU processing of the CNN.
The post-processing mainly includes pulling from the CNN output layer and formatting the data into a OpenCV grayscale matrix.
Our navigation algorithm uses the generated mask by the CNN to determine a path for the vehicle to travel, which operates in parallel to the other ROS nodes.
Figure \ref{baseline_latency_graph} shows this timing breakdown for the fastest network that we tested in ENet.

As CNN is the most compute-intensive function, we expected forward propagation time dominates the overall latency. However, surprisingly, post-processing stage took almost as long of time as the actual forward-propagation. To find the root cause of this long latency, we dug the code of Caffe and NVCaffe and found that the last layer if CNN is not well parallelized in both frameworks. Though the last layer requires a high-level parallelism because it traverses all the pixels to find the proper label for prediction, the frameworks run them on a CPU. One reason may be to reduce memory copy latency by running the last layer on CPU, which can directly serve the final prediction in CPU side system memory. This optimization may work for server systems that GPU uses a separate device memory. But, in the mobile systems that use a unified shared memory, the CPU-side computation only incur performance overhead. We provide more details about this and propose an optimization in Section~\ref{sec:optimization}.

\textit{\textbf{Observation 3:} The number of layers and the amount of parameters have strong impact to the overall throughput. Though more layers and parameters help increase accuracy for complex objects, they need to be trimmed if throughput is the first-priority design consideration, which is true for the systems that need real-time performance.}

\textit{\textbf{Observation 4:} From the end-to-end latency breakdown, the CNN post-processing needs to be carefully handled or else it can significantly extend the end-to-end latency. Many open-source CNN benchmarks and reference codes commonly overlook or ignore this important step, especially because most of the frameworks are designed for server systems that has different CPU-GPU organizations than mobile systems.}


%% file: energy_efficiency.tex
\subsection{Energy Efficiency}

\begin{figure}[htp]
    \centering
    \includegraphics[width=8cm]{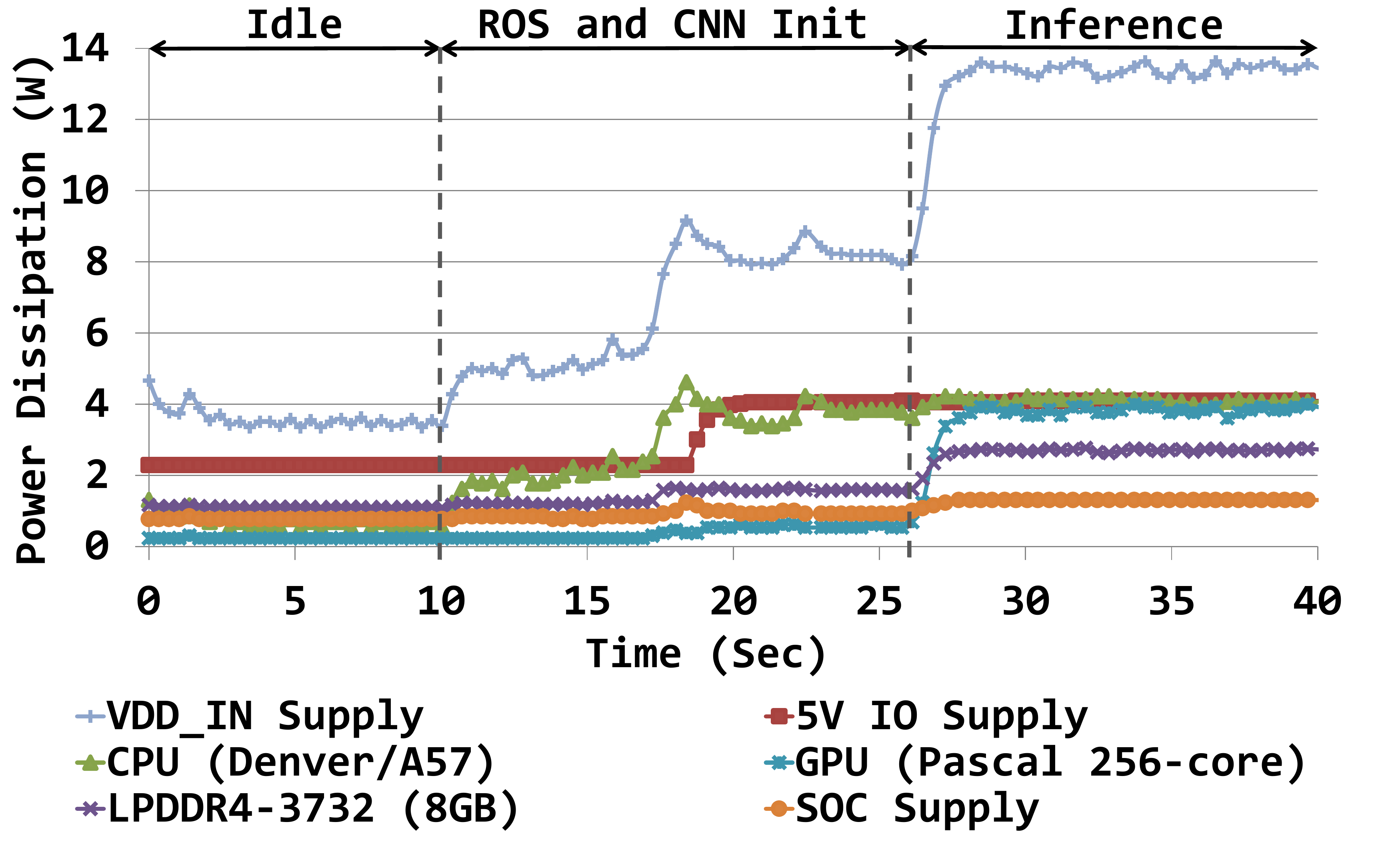}
    \caption{Power monitoring sequence with ENet deployed on the TX2. }
    \label{power_init_graph}
\end{figure}

To understand the energy efficiency of the models with our system, we break down the power consumption of each component, from when the vehicle is in an idle state and until it is steadily active while performing inference.
We conduct measurements while reading from six different INA3221 voltage/current sensors on the Jetson, approximately every 300ms. 
Figure \ref{power_init_graph} shows this sequence and the associated power dissipation with ENet deployed on the system.
The 5V rail powers the ZED camera, USB-Ethernet Adapter for Lidar communication, GPS, Compass, USB hub, and the Teensy uC (in descending order), all of which consume a relatively constant amount of power in our design.
CPU handles navigation control along with ROS node management while GPU is dedicated towards CNN processing which operates in parallel to the other ROS nodes.
Both Jetson platforms are designed with unified memory, where the LPDDR4 is shared between GPU and CPU.
The SOC is responsible for a few background operations specific to the Jetson design, such as managing the BPMP (Boot and Power Management Processor) and IRAM.

\begin{figure}[htp]
    \centering
    \includegraphics[width=8cm]{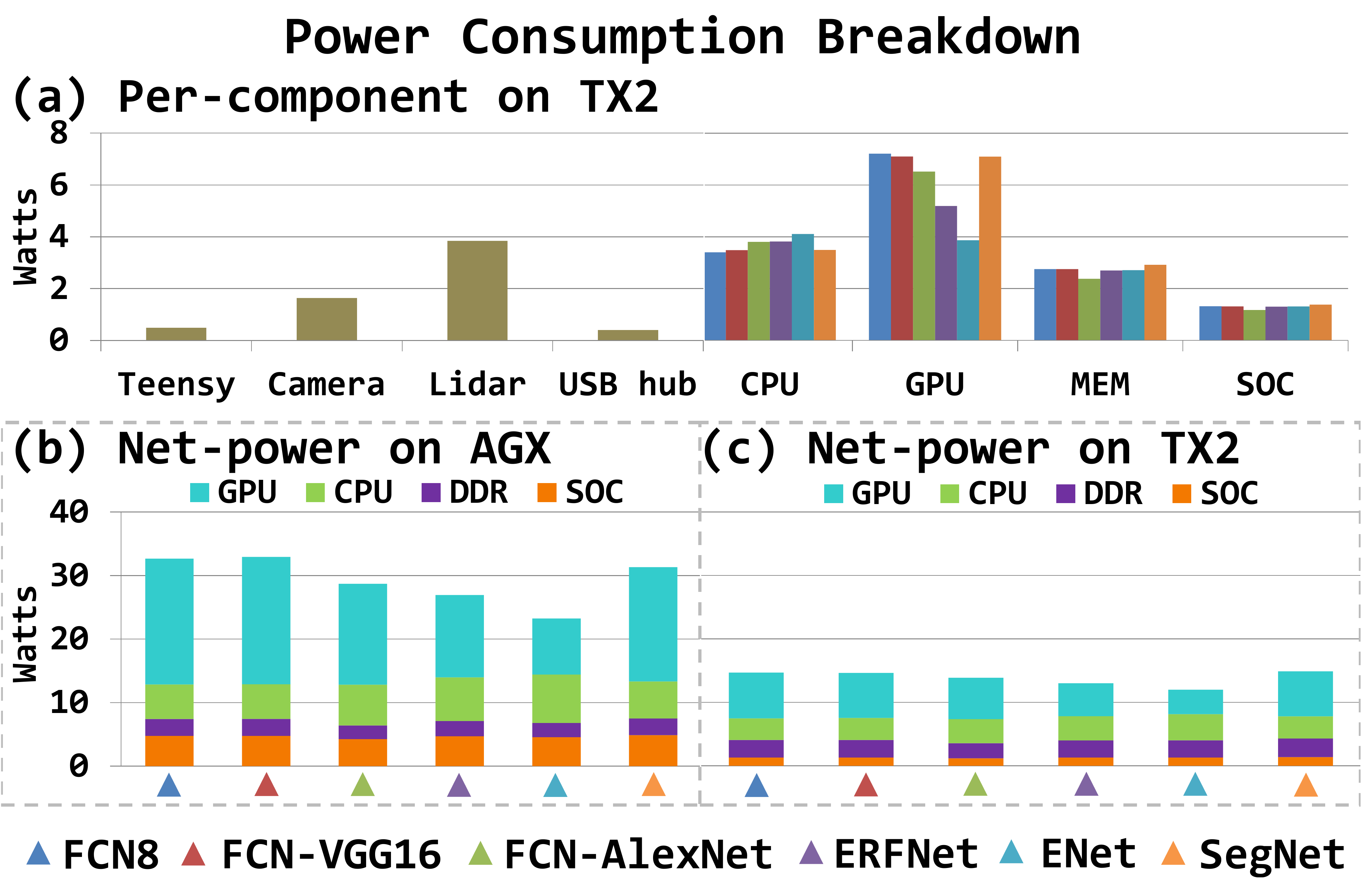}
    \caption{System and CNN power dissipation (a) Components on the TX2 (b) AGX Xavier accumulation (c) TX2 accumulation }
    \label{per_net_power_graph}
\end{figure}

Because CNN is always executing on GPU, the CPU-based navigation control is limited by the throughput of the CNN. This GPU-to-CPU dependency raises the question as to how the CNN speed affects the power consumption of GPU and CPU at the system-level. In Figure \ref{per_net_power_graph}, we breakdown the power of each component and CNN model, as well as the accumulated system-level power. Different colors indicate different models. As can be seen in Figure~\ref{per_net_power_graph}(a), CPU and GPU dominates power consumption, though there are some differences among the models especially in GPU power. However, one notable finding from this per-component power consumption is that a Lidar consumes as much power as the two processors and even more than CPU. 

\begin{figure}[htp]
    \centering
    \includegraphics[width=8cm]{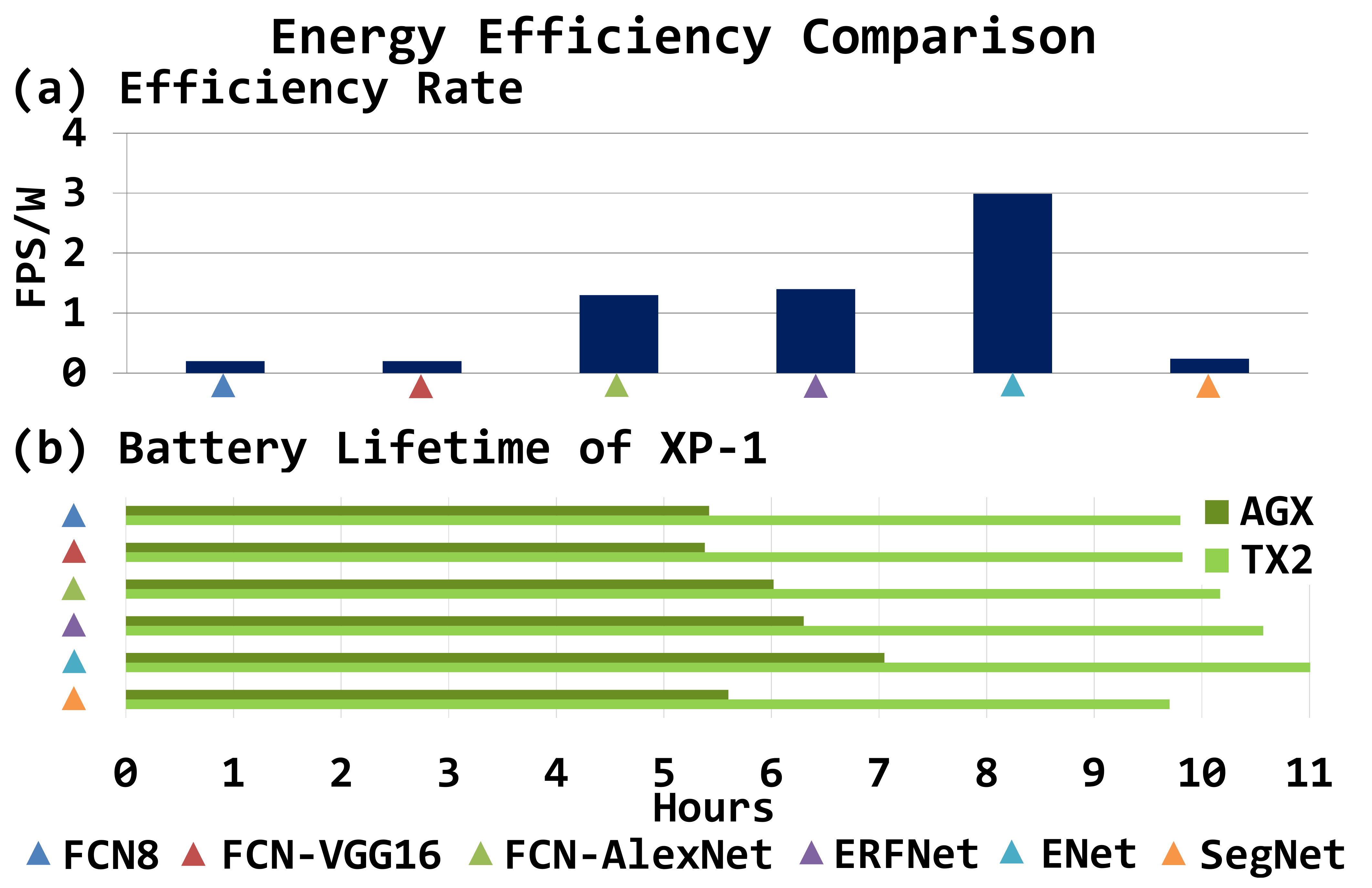}
    \caption{Energy efficiency comparison of the CNN models in terms of: (a) Frames per second per watt (b) Maximum battery lifetime of the XP-1}
    \label{energy_eff_graphs}
\end{figure}

Considering the net power consumption with the models, power was generally lower as the model speed increased. We found an exception to this trend when comparing FCN-AlexNet and ERFNet. FCN-AlexNet was slightly faster, however, it consumed about 1.4W more from GPU than ERFNet. We then dug deeper to try and understand the efficiency rates of the models, which we measure and report in Figure \ref{energy_eff_graphs}, along with the maximum battery lifetime of our system comparing each model and platform.

\textit{\textbf{Observations 5:} From the power consumption measurements with ENet, the Lidar sensor (including all of its components) consumes a very comparable amount of power compared to the GPU chip alone.}

\textit{\textbf{Observations 6:} From the energy efficiency comparisons, we find that model speed is a good indicator of its energy efficiency, but not necessarily a direct correlation.}

%% file: optimizations.tex
\section{Optimizations}\label{sec:optimization}
\subsection{Acceleration of \textit{argmax} layer on GPU} 

\begin{figure}[htp]
    \centering
    \includegraphics[width=6cm]{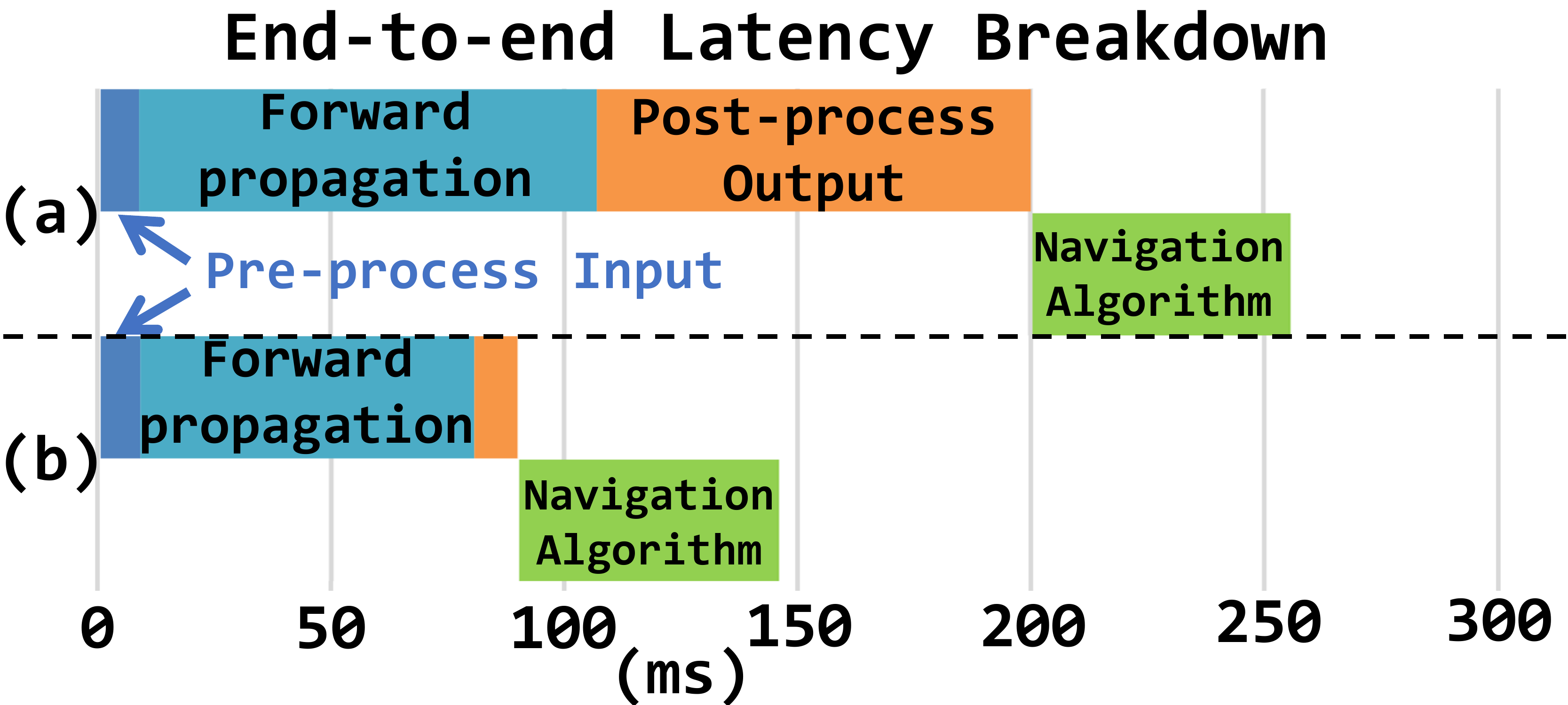}
    \caption{Optimized End-to-end latency breakdown with ENet deployed on the TX2. (a) Original latency using CPU-based argmax (b) Optimized latency using GPU-based argmax}
    \label{argmax_latency_breakdown}
\end{figure}

We found an opportunity for Caffe \cite{caffe} and NVCaffe \cite{nvcaffe} to be optimized, which was also discovered by \cite{enet_argmax} in 2018, but unaddressed by either Caffe developer, up until at least this paper.
From the statistics of the design mentioned in the Performance section, we found a significant bottleneck while the network output was being post-processed, which turned out to be when calculating the argmax of the CNN’s final layer. During the argmax stage, all of the possibly predicted classes are iterated through for each pixel, and the index consisting of the largest value contains the final prediction for that pixel. From digging through the source code of Caffe, we found that this layer was being processed by the CPU, which computes the argmax primarily sequential for each pixel. Given that our network output mask was 480x360 pixels, this adds up to over 1.7M sequential operations. 
Since the argmax operation for semantic segmentation does not have dependencies among neighboring pixels, each pixel does not need to be checked sequentially. By fully parallelizing the pixel-wise argmax towards the GPU, we can allocate one CUDA thread per pixel to achieve a significantly higher performance improvement. 
Figure \ref{argmax_latency_breakdown} shows the end-to-end timing breakdown of our system comparing both CPU-based and GPU-based argmax implementations.

\begin{figure}[htp]
    \vspace*{0.15cm}
    \centering
    \includegraphics[width=8cm]{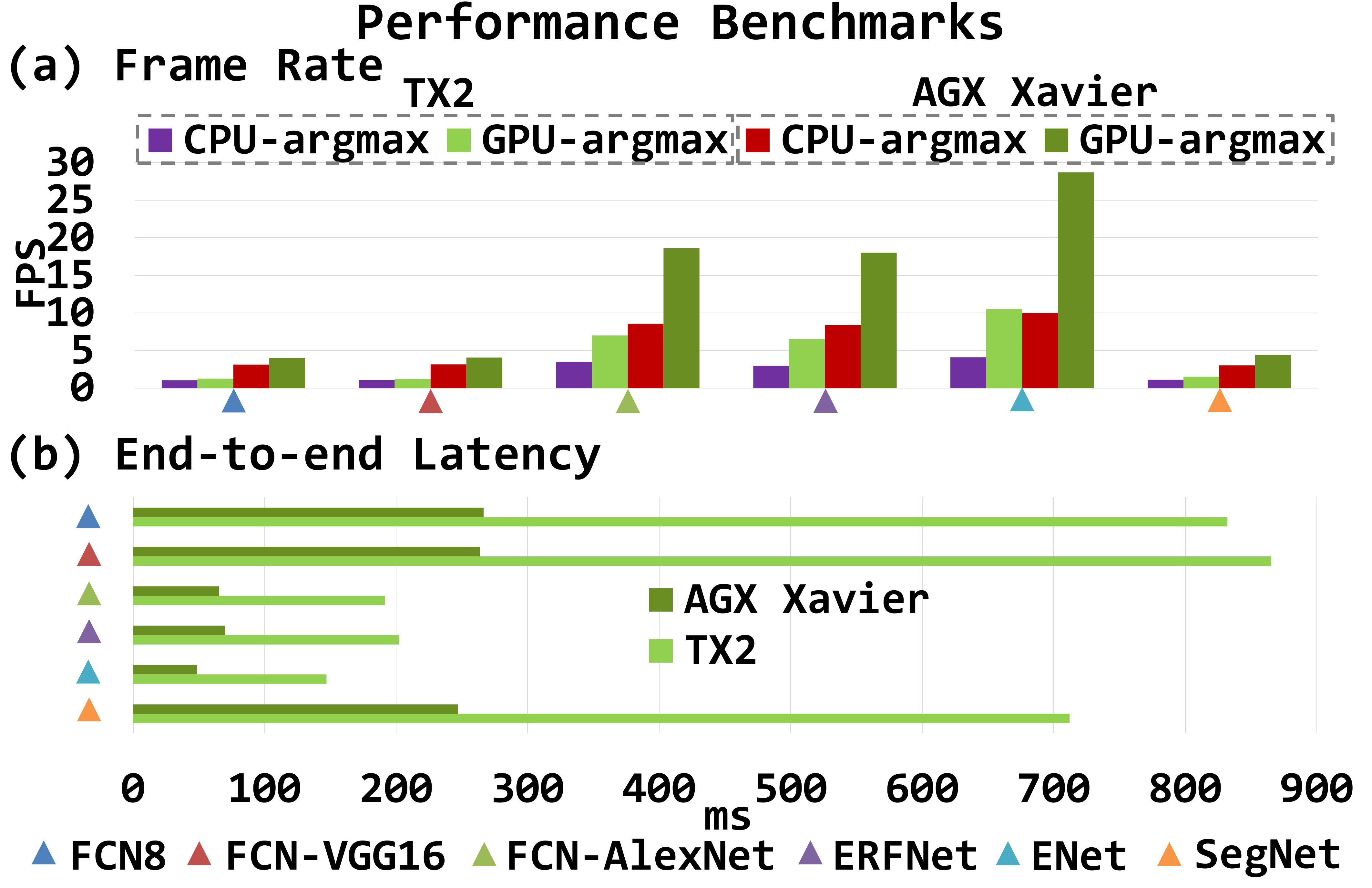}
    \caption{Optimized performance with the CNN models (a) FPS (b) Optimized End-to-end latency. }
    \label{optimized_performance_graph}
\end{figure}

Additional performance results of the various models comparing the CPU-based argmax with the GPU-based argmax are shown in Fig. \ref{optimized_performance_graph}a, when tested on our Jetson platforms with CUDA 10 and cuDNN 7.5. 
Fig. \ref{optimized_performance_graph}b shows the resulting end-to-end latency on our Jetson boards with our optimized GPU-based argmax implementation.
Our mask-based local navigation algorithm accounts for about 56ms of the reported TX2 latencies and about 22ms of the AGX Xavier latencies.

We also experimented the models with TensorRT 7.1 (latest compatible with JetPack, at the time of this writing) and found that they didn't support the argmax layer. However, from experimenting with their reference code, the results indicated that they computed the argmax via CPU while using additional host-side code.

\subsection{Boosting FPS with Sensor Fusion} 

\begin{figure}[htp]
    \centering
    \includegraphics[width=4.675cm]{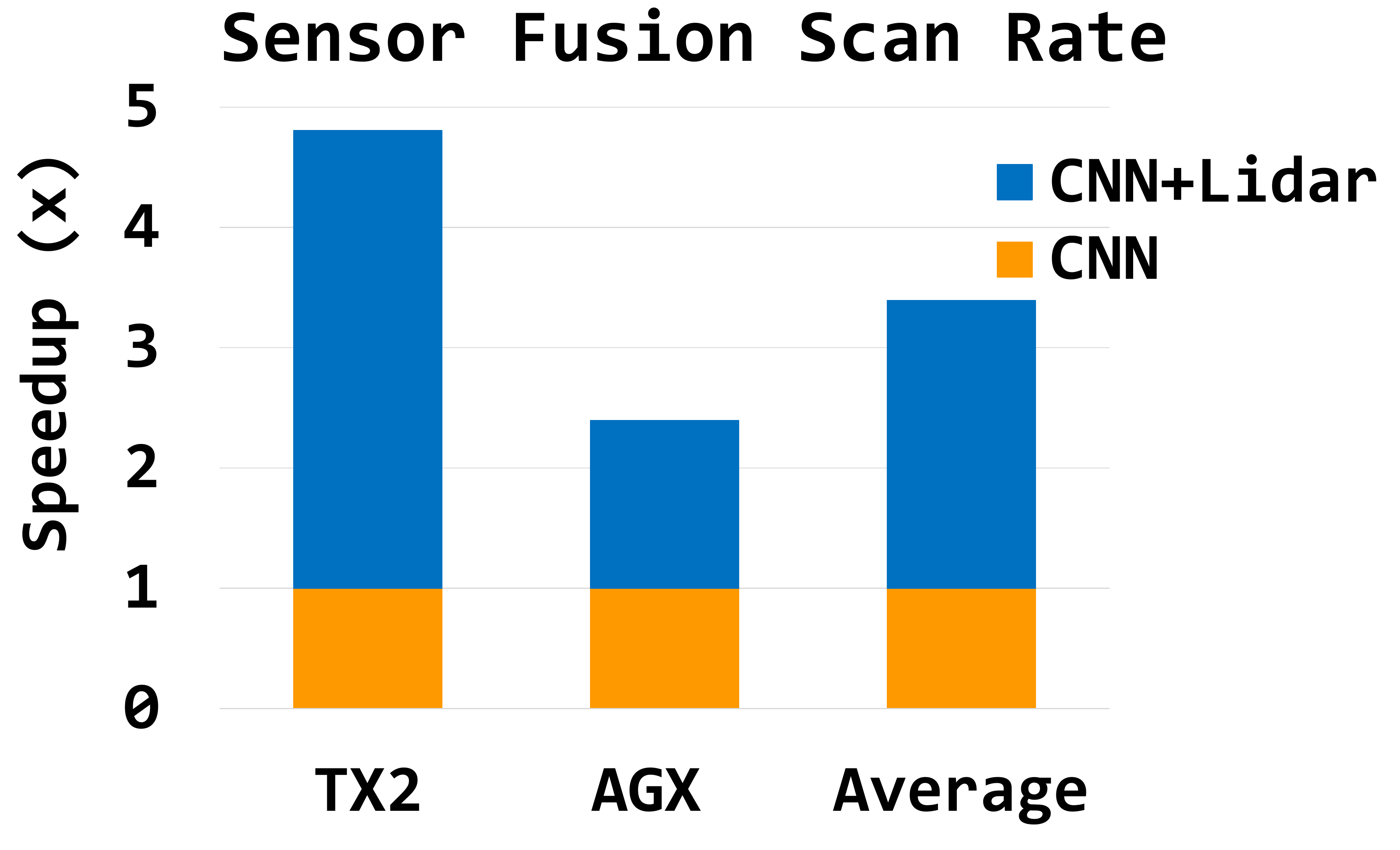}
    \caption{Sensor fusion scan rate speedup when fusing our camera (using ENet) with Lidar. }
    \label{cnn_lidar_graph}
\end{figure}

To continue accelerating the system scan rate and ensure rapid updates to our vehicle, we incorporate a sensor fusion by using the segmented mask from our camera and our Lidar sensor.
Fusing both sensors together enables performance that meets the real-time frame rate standard of 30 frames per second.
Figure \ref{cnn_lidar_graph} shows the scan rate speedup from fusing lidar with our camera-based vision.
This results in a system response time of at most 27ms on both TX2 and AGX Xavier configurations.

%% file: analysis.tex
\section{Analysis and Discussions}

\begin{figure}[htp]
    \vspace*{0.15cm}
    \centering
    \includegraphics[width=6cm]{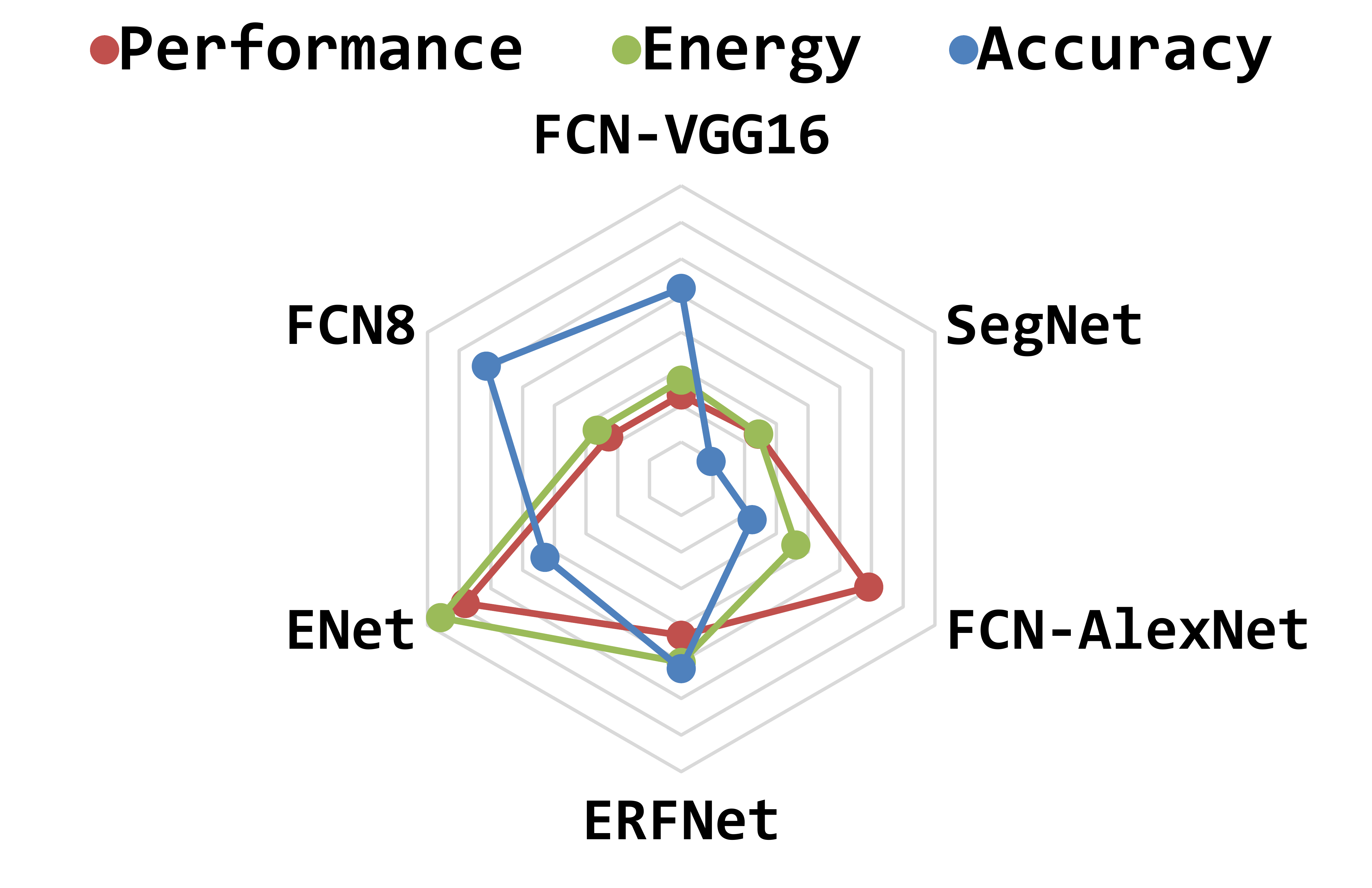}
    \caption{Normalized comparison between the models (points closer to the CNN name-label are better).}
    \label{acc_perf_energy_graph}
\end{figure}

Considering the accuracy, performance, and energy efficiency tradeoffs of our systems comparing the different CNNs, an optimal model can be correctly selected. In Figure \ref{acc_perf_energy_graph}, each is equally weighted and normalized for a fair comparison between models.
Normalization is based on the mean and standard deviation of mIoU (accuracy), optimized forward-pass time (performance), and net power consumption (energy-efficiency).

Based on all of the statistics and results, we can recommend ENet for the mobile-edge vehicle as it achieves more optimal performance and energy-efficiency compared to the other CNNs, with a relatively minor reduction in accuracy. If the accuracy for camera-based vision needs to be improved, ERFNet would be our next recommendation as long as the speed of camera-based vision can afford to be reduced by roughly 40\%, which may be acceptable for slower speed robots. If accuracy is the highest priority while performance and energy efficiency can be ignored, FCN8 and FCN-VGG16 would be the best solutions of these models. Based on our qualitative assessment, we found that FCN8 (stride of 8) is better for segmenting the contours of the obstacles while FCN-VGG16 (stride of 32) is more so suited for generic clustering of obstacles, meaning that their edges and contours would not be as important. For navigation in earthquake-struck zones, FCN8 would be better than FCN-VGG16 due to the higher priority of needing to detect the irregular contours of the road cracks and other irregular types of obstacles encountered.  

To further understand why some of the models do better than others and like we discussed in the earlier sections, we broke each model down at a per-layer level. The more accurate models compute over 100M parameters, although they suffer from worse performance and energy efficiency. The models originally designed for semantic segmentation (ERFNet, ENet, SegNet) show that by widely distributing the parameters across layers/modules, they can be designed efficiently with a low number of total parameters. For instance, SegNet computes fewer parameters than the models converted to FCN because the SegNet decoder reuses max-pooling indices from the encoder when upsampling the feature maps. However, as the accuracy results showed, this technique turned out to be detrimental to SegNet's accuracy, resulting in a worse accuracy than all the other models we tested.
ENet computes the fewest amount of parameters because of its heavy utilization of bottleneck modules based on dilated and asymmetric convolutions, which significantly reduces the size and execution time of each convolution, as shown in the graphs mentioned earlier. 
These design features in ENet resulted in it as performing better than all the other models we tested, in terms of both FPS, latency, as well as energy-efficiency, all while not suffering from significant tradeoffs in terms of the accuracy, making it an ideal model for self-navigation on mobile platforms.

%% file: conclusion.tex
\section{CONCLUSION}

In this paper, we proposed a mobile-based unmanned vehicle design that operates in real-time and can detect unique obstacles in earthquake-struck zones amongst other areas where road conditions are not ideal.
Our proof-of-concept vehicle design shows an example as to how our earthquake-site image database can be used to help vehicles autonomously navigate in these types of conditions with irregular obstacles.
Our experiments demonstrate that the characterized system and proposed image database performs well it terms of detection capability while meeting real-time operating constraints.
We also highlighted key design and system considerations that need to be made when developing and deploying FCN models for mobile-based self navigation.
By considering our design, characterization, optimizations, analysis, and annotated image database, system integrators can better plan their self-navigation platforms that navigate based on semantic segmentation with the added capability of detecting irregular objects that would exist at sites where road conditions may be non-ideal or even damaged, such as in an earthquake-struck zone.


%% file: bibliography.tex






\renewcommand*{\bibfont}{\footnotesize}
\printbibliography